# Personal Universes:
## A Solution to the Multi-Agent Value Alignment Problem

**Roman V. Yampolskiy**
Computer Engineering and Computer Science
Speed School of Engineering
University of Louisville
roman.yampolskiy@louisville.edu

**Abstract**
AI Safety researchers attempting to align values of highly capable intelligent systems with those of humanity face a number of challenges including personal value extraction, multi-agent value merger and finally in-silico encoding. State-of-the-art research in value alignment shows difficulties in every stage in this process, but merger of incompatible preferences is a particularly difficult challenge to overcome. In this paper we assume that the value extraction problem will be solved and propose a possible way to implement an AI solution which optimally aligns with individual preferences of each user. We conclude by analyzing benefits and limitations of the proposed approach.

## 1. Introduction to the Multi-Agent Value Alignment Problem

Since the birth of the field of Artificial Intelligence (AI) researchers worked on creating ever capable machines, but with recent success in multiple subdomains of AI [1-7] safety and security of such systems and predicted future superintelligences [8, 9] has become paramount [10, 11]. While many diverse safety mechanisms are being investigated [12, 13], the ultimate goal is to align AI with goals, values and preferences of its users which is likely to include all of humanity.

Value alignment problem [14], can be decomposed into three sub-problems, namely: personal value extraction from individual persons, combination of such personal preferences in a way, which is acceptable to all, and finally production of an intelligent system, which implements combined values of humanity.

A number of approaches for extracting values [15-17] from people have been investigated, including inverse reinforcement learning [18, 19], brain scanning [20], value learning from literature [21], and understanding of human cognitive limitations [22]. Assessment of potential for success for particular techniques of value extraction is beyond the scope of this paper and we simply assume that one of the current methods, their combination, or some future approach will allow us to accurately learn values of given people. Likewise, we will not directly address how, once learned, such values can be represented/encoded in computer systems for storage and processing. These assumptions free us from having to worry about safety problems with misaligned AIs such as perverse instantiation or wireheading [23], among many others [24].

The second step in the process requires an algorithm for value aggregation from some and perhaps even all people to assure that the developed AI is beneficial to the humanity as a whole. Some have suggest that interests of future people [25], potential people [26] and of non-human animals and other sentient beings, be likewise included in our "Coherent Extrapolated Volition" (CEV) [27], which we would like superintelligent AI to eventually implement. However, work done by moral philosophers over hundreds of years indicates that our moral preferences are not only difficult to distil in a coherent manner (anti-codifiability thesis) [28], they are also likely impossible to merge without sacrificing interests of some people [29, 30], we can say it is the Hard problem of value alignment. Results from research into multivariate optimization and voting based preference aggregation support similar conclusions [31-33].

Perhaps we should stop trying to make "one size fits all" approach to the optimization of the universe work and instead look at potential for delivering an experience customized to individual users. The superintelligent systems we are hoping to one day create, with the goal of improving lives of all, may work best if instead they strive to optimize their alignment with individual lives of each and every one of us, while giving us all freedom to be ourselves without infringing on preferences of other sentient [34, 35] beings. Such a system due to its lower overall complexity should also be easier to design, implement and safeguard.

## 2. Individual Simulated Universes

It has been suggested that future technology will permit design [36] and instantiation of high fidelity simulated universes [37-41] for research and entertainment ([42], chapter 5) purposes as well as for testing advanced AIs [43-46]. Existing work and recent breakthroughs in virtual reality, augmented reality, inter-reality, haptics, and artificial consciousness combined with tremendous popularity of multiplayer virtual worlds such as Second Life [47-49] or Ultima Online [50] provide encouraging evidence for the plausibility of realistic simulations.

We can foresee, in a not so distant future, a point at which visual and audio fidelity of the simulations, as well as for all other senses [51] becomes so high that it will not be possible to distinguish if you are in a base reality or in a simulated world, frequently referred as hyperreality [52, 53]. In principle, it should be possible to improve local fidelity (measurable by the agent) of the simulated reality to levels beyond base reality, for example to the point of more precise measurements being possible with special instrumentation. This would effectively reverse the resolution relationship between the two realities making the base reality less believable on local scale. A variant of a Total Turing Test [54, 55], we shall call a Universal Turing Test (UTT) could be administered in which the user tries to determine if the current environment is synthetic or not [56] even if it is complex enough to include the whole universe, all other beings (as philosophical zombies [57]/Non-Playing Characters (NPCs)) and AIs. Once the UTT is consistently passed we will know, the hyperreality is upon us.

Consequently, we suggest that instead of trying to agree on convergent, universal, diverse, mutually beneficial, equalizing, representative, unbiased, timeless, acceptable to all, etc. moral/ethical norms and values, predicated on compromise [58], we look at an obvious alternative.

Specifically, we suggest that superintelligent AIs should be implemented to act as personalized simulations - *Individual Simulated Universes* (ISU) representing customized synthetically generated [7, 59] mega-environments, in the "a universe per person multi-verse framework", which are optimally and dynamically adjusting to align their values and preferences to the Personal CEV [60] of sentient agents calling such universes "home".

Aaronson describes the general idea as "… an infinite number of sentient beings living in simulated paradises of their own choosing, racking up an infinite amount of utility. If such a being wants challenge and adventure, then challenge and adventure is what it gets; if nonstop sex, then nonstop sex; if a proof of P≠NP, then a proof of P≠NP. (Or the being could choose all three: it's utopia, after all!)" [61]. Bostrom estimates that our galactic supercluster has enough energy to support trillions of such efficiently [62] simulated universes [63]. Features of related phenomenon have been described in literature as [64]: dematerialization [65], ephemeralization [66], time-space compression [67], miniaturization [68], densification [69], virtualization [70], digitization [71], and simulation [72].

Faggella talks about opportunities presented in the virtual world over what is possible in the present reality [73]: "… 'freedom' could only extend so far in a real world as to border on impinging on the 'freedom' of others. Complete freedom would imply control over one's environment and free choice to do what one would chose with it. It seems easy to understand how this might imply the threatening of the freedom of others in the same physical world. … Not to mention, the physical world has many impinging qualities that would hinder any semblance of complete freedom. Matter has qualities, light has qualities, and physical bodies (no matter how enhanced) will always have limitations. If you'd like to change an aspect of our character or emotional experience, for example, we'd have to potentially tinker with brain chemicals … . In a virtual reality, we are potentially presented not only with the freedom to extend beyond physical limitations (to transport to different times or places, to live within self-created fantasy worlds, to eliminate death and any physical risk), we would also be granted freedom from impinging or effecting others – and so allow for their full freedom an a separate virtual reality as well. … For this reason, it seems to make sense that … we might encounter a Bostrom-like 'Singleton' to rule the physical world, and a great sea of individual consciousnesses in the virtual world. The 'Singleton' could keep our computational substrates safe from harm and eliminate competition or danger in the physical world, while our virtual 'selves' would be capable of expressing and exploring the epitome of freedom on our own terms in a limitless virtual world of our own creation."

This means that an ISU can be anything a user truly wishes it to be including dangerous, adversarial, competitive, and challenging at all levels of user competence like levels in a well-designed video game. It will let a user be anything they want to be including a malevolent actor [74, 75], a privileged person (like a king) or the exact opposite (a slave), or perhaps just a selfish user in an altruistic universe. A personalized universe doesn't have to be fair, or just or free of perceived suffering and pain [76]. It could be just a sequence of temporary fantasies and hopefully what happens in your personalized universe stays in your personalized universe. ISU's goal is to cater to the world's smallest minority and its preferences, you [77, 78]! Moreover, the good news is that we know that we are not going to run out of Fun [79] even if we live much longer lives [80].

If an agent controlling the environment is not well aligning with a particular individual for whom the environment is created (during early stages of development of this technology) it may be necessary to use precise language to express what the user wants. The now defunct Open-Source Wish Project (OSWP) [81] attempted to formulate in precise and safe form such common wishes as: immortality, happiness, omniscience, being rich, having true love, omnipotence, etc [23].

For example the latest version of the properly formed request for immortality was formalized as follows: "I wish to live in the locations of my choice, in a physically healthy, uninjured, and apparently normal version of my current body containing my current mental state, a body which will heal from all injuries at a rate three sigmas faster than the average given the medical technology available to me, and which will be protected from any diseases, injuries or illnesses causing disability, pain, or degraded functionality or any sense, organ, or bodily function for more than ten days consecutively or fifteen days in any year; at any time I may rejuvenate my body to a younger age, by saying a phrase matching this pattern five times without interruption, and with conscious intent: 'I wish to be age,' followed by a number between one and two hundred, followed by 'years old,' at which point the pattern ends - after saying a phrase matching that pattern, my body will revert to an age matching the number of years I started and I will commence to age normally from that stage, with all of my memories intact; at any time I may die, by saying five times without interruption, and with conscious intent, 'I wish to be dead'; the terms 'year' and 'day' in this wish shall be interpreted as the ISO standard definitions of the Earth year and day as of 2006. [81]" Of course, this is still far from foolproof and is likely to lead to some undesirable situations, which could be avoided by development of a well-aligned system.

## 3. Benefits and Shortcomings of Personalized Universes

ISUs can be implemented in a number of ways, either by having perfect emulations of agents reside in the simulated universe or by having current biological agents experience fully realistic simulated environments (while robotic systems take care of their bodies' biological needs), see Faggella's review of possible variants of virtual reality [82]. Both options have certain desirable properties, for example, software versions of users are much easier to modify, reset to earlier memory states [83], upgrade and backup [84, 85], while biological agents are likely to have stronger identity continuity [86]. Emulations can also be taken as snapshots from different points in the person's life and set to exist in their own independent simulations multiplying possible experiences [34] for the subset of agents derived from that particular individual. In both virtual and uploaded scenarios, it is probably desirable for the user to "forget" that they are not in the base reality via some technological means with the goal of avoiding Solipsism syndrome[1].

Our proposal doesn't just allow us to bypass having to find a difficult to compute approximation to a likely impossible to solve problem of multi-agent value aggregation, but it also provides for a much better "customer experience" free of compromise on even small details which may be important to that individual. Additionally, virtual existence makes it possible to have an "undo button" for actions/experiences user might regret, something not always possible in the world of physical reality. Last, but not least any existential risks related to this particular AIs failure are limited to the simulated universe and its virtual inhabitants, not to the humanity and all life forms.

---

[1] https://en.wikipedia.org/wiki/Solipsism_syndrome

Of course, like any AI safety mechanism ours has certain weaknesses, which will have to be explicitly addressed. Those include having to withstand agents with extreme preferences, who may wish to prevent others from exercising their self-determination and may attempt to hack and sabotage ISUs or even base reality (which should be easier to secure, with most agents and their complex preferences out of the way). Another area of concern is problems with superintelligence serving as "operating system" for the base reality and allocating non-conflicting resources for the ISUs. Finally, we should study how the philosophical questions of living in a "fake" world vs "real" world, even if it is not possible to distinguish between them by any means, impacts human psychology and well-being.

It is also important to figure out a metric to measure user-relative quality of the simulation experience not just from fidelity point of view but also from users overall satisfaction with how their values, goals and preferences are being serviced, such metrics are notoriously hard to design and easy to abuse [87]. Potential ideas may include user feedback both from within the simulation and while outside observing a recording of themselves in the simulation, feedback after trying other simulations and potentially all other simulations, and peer-review from other conscious agents both from outside and from within the same environment.

It is possible to let users "play" in other's universes and perhaps as other characters and to allow them to discover and integrate new values to which their universe will dynamically adopt. It may also be possible for two or more agents to decide to cohabit the same universe by coming to accept a mutually satisfying set of values, but of course their individual alignment with the environment would be reduced and so it is important to provide them with a "divorce" option. We are assuming a well aligned AI, which will not attempt to directly hack the agent to game the feedback score, but out of caution, we do not recommend evolutionary competition [88-90] between ISUs as that can lead to adversarial behaviors between superintelligent agents even the base reality superintelligence would not be able to resolve.

## 4. Conclusions

In this exploratory paper, we advocated a solution to the hardest of the three subproblems of multi-agent value alignment, specifically value aggregation. Our "in the box" solution suggests replacing one-size-fits-all model of value satisfaction with customized and highly optimized approach which is strictly superior for all possible agents not valuing decreasing quality of value alignment for other agents. Some existing evidence from cosmology may be seen as suggesting that perhaps this approach is not so novel and in fact has already been implemented by earlier civilizations, and this universe is already a part of a multiverse [91, 92] generated by intelligence [93]. While some significant concerns with the philosophical [94], social [95] and security [96, 97] problems associated with personalized universes remain, particularly with regards to securing base reality, the proposal has a number of previously described advantages. Such advantages are likely to make it attractive to many users or to at least be integrated as a part of a more complex hybrid solution scheme. The decisions made by users of personal universes are also a goldmine of valuable data both for assessment of agents and for providing additional data to improve overall AI alignment [98]. We will leave proposals for assuring safety and security of cyberinfrastructure running personalized universes for future work. The main point of this paper is that a personal universe is a place where virtually everyone can be happy.

## Acknowledgments

The author is grateful to Elon Musk and the Future of Life Institute and to Jaan Tallinn and Effective Altruism Ventures for partially funding his work on AI Safety. Special thank you goes to all NPCs in this universe.